\documentclass[letterpaper, 10 pt, conference]{ieeeconf}  %
\IEEEoverridecommandlockouts
\usepackage{amsmath,amsfonts}
\usepackage{array}
\usepackage[caption=false,font=normalsize,labelfont=sf,textfont=sf]{subfig}
\usepackage{textcomp}
\usepackage{stfloats}
\usepackage{url}
\usepackage{verbatim}
\let\labelindent\relax  
\usepackage{enumitem}
\usepackage{titlesec}
\usepackage{graphicx}
\usepackage[ruled,vlined]{algorithm2e}
\usepackage[switch]{lineno}
\usepackage{color}
\usepackage{hyperref} 
\usepackage{booktabs}
\usepackage{multirow}
\def\BibTeX{{\rm B\kern-.05em{\sc i\kern-.025em b}\kern-.08em
    T\kern-.1667em\lower.7ex\hbox{E}\kern-.125emX}}
\usepackage{balance}

\usepackage{xspace}

\begin{document}
\title{\textbf{L}anguage-\textbf{D}riven \textbf{P}olicy \textbf{D}istillation for Cooperative Driving in Multi-Agent Reinforcement Learning}

\author{
Jiaqi~Liu, Chengkai~Xu, Peng~Hang, Jian Sun, Wei Zhan, Masayoshi Tomizuka, and Mingyu Ding
\thanks{Jiaqi Liu and Mingyu Ding are with the Department of Computer Science at University of North Carolina at Chapel Hill, NC, USA.\{jqliu,md\}@cs.unc.edu}
\thanks{Chengkai Xu, Peng Hang and Jian Sun are with the College of Transportation and Key Laboratory of Road and Traffic Engineering, Ministry of Education, Tongji University, Shanghai 201804, China.}
\thanks{Masayoshi Tomizuka and Wei Zhan are with the Department of Mechanical Engineering at the University of California, Berkeley, CA, USA.\{tomizuka,wzhan\}@berkeley.edu}
\thanks{Corresponding author: Mingyu Ding}
}


\maketitle

\begin{abstract}
The cooperative driving technology of Connected and Autonomous Vehicles (CAVs) is crucial for improving the efficiency and safety of transportation systems. Learning-based methods, such as Multi-Agent Reinforcement Learning (MARL), have demonstrated strong capabilities in cooperative decision-making tasks. However, existing MARL approaches still face challenges in terms of learning efficiency and performance. In recent years, Large Language Models (LLMs) have rapidly advanced and shown remarkable abilities in various sequential decision-making tasks. To enhance the learning capabilities of cooperative agents while ensuring decision-making efficiency and cost-effectiveness, we propose LDPD, a language-driven policy distillation method for guiding MARL exploration. In this framework, a teacher agent based on LLM trains smaller student agents to achieve cooperative decision-making through its own decision-making demonstrations. The teacher agent enhances the observation information of CAVs and utilizes LLMs to perform complex cooperative decision-making reasoning, which also leverages carefully designed decision-making tools to achieve expert-level decisions, providing high-quality teaching experiences. The student agent then refines the teacher’s prior knowledge into its own model through gradient policy updates. The experiments demonstrate that the students can rapidly improve their capabilities with minimal guidance from the teacher and eventually surpass the teacher's performance. Extensive experiments show that our approach demonstrates better performance and learning efficiency compared to baseline methods.
\end{abstract}


\section{Introduction}
Recent advancements in Connected and Autonomous Vehicle (CAV) technology have enabled vehicles to communicate and collaborate, revolutionizing the transportation industry by enhancing overall traffic management~\cite{hang2022decision,liu2024delay,qu2023advancements}. Cooperative driving technology is a critical component of CAV deployment, which enables them to make efficient decisions in dynamic environments, thereby enhancing safety and traffic flow. However, despite the significant potential of cooperative decision-making technology, current methods still face numerous challenges in terms of scenario applicability, decision-making efficiency, and safety, making large-scale real-world deployment a distant goal~\cite{hang2021cooperative}.

The rapid development in deep learning has made learning-based approaches, such as Multi-Agent Reinforcement Learning (MARL), effective for cooperative decision-making~\cite{zhang2023coordinating}. In MARL, each agent explores strategies to maximize its own reward, thereby identifying optimal decision-making policies of CAV, which has been employed by many researchers~\cite{liu2024cooperative,chen2023deep,zhang2024multi}. 
Despite the notable advantages of MARL, challenges remain, such as difficulties in algorithm convergence, low exploration efficiency, and high sampling costs~\cite{zhang2024multi}. On the other hand, recent advancements in Large Language Models (LLMs) have demonstrated impressive zero-shot generalization and complex logical reasoning capabilities across various decision-making tasks~\cite{zhang2024towards}. These capabilities are particularly well-suited for assessing complex situations in human traffic environments and achieving coordinated control between vehicles and traffic systems. A few studies have explored the integration of LLMs into traffic and vehicle systems to enhance the safety and efficiency of transportation networks~\cite{fang2024codrivingllm,hu2024agentscodriver}. 
However, in real-time, dynamic traffic scenarios, existing LLMs struggle to enable CAVs to make rapid and efficient decisions. Additionally, the high deployment costs hinder large-scale implementation, significantly limiting the potential of LLMs~\cite{zhou2023large}.

To address these challenges, we propose a novel approach that leverages the reasoning efficiency of MARL-based agents alongside the extensive world knowledge of LLMs. This approach involves designing a teacher-student policy distillation framework to facilitate the efficient transfer of LLMs' world knowledge to MARL agents.
we present LDPD, a language-driven policy distillation framework designed to guide MARL exploration. The framework features a teacher agent powered by LLM and multiple student agents (CAVs) modeled with MARL, enabling effective knowledge transfer and collaborative learning.
During the training phase, the teacher agent receives observation information from students, leveraging its powerful zero-shot and reasoning capabilities to make expert-level decisions. It then guides the smaller student agents through its actions, enabling them to perform efficient sampling and policy learning. After the learning phase, these Student policies will function as the controllers for CAVs, making decisions independently of the teacher during deployment.
For teacher agent,  the Observation module preprocesses and enhances student observations, which are then sent to the LLM-based Planner for decision-making. The Planner integrates the data using agent tools, and expert-level outputs are generated. These expert decisions are used to train the student policies. The student agents are trained in a decentralized MARL framework, using the Actor-Critic method to update their action policies.

\begin{figure}
  \centering
  \includegraphics[width=0.40\textwidth]{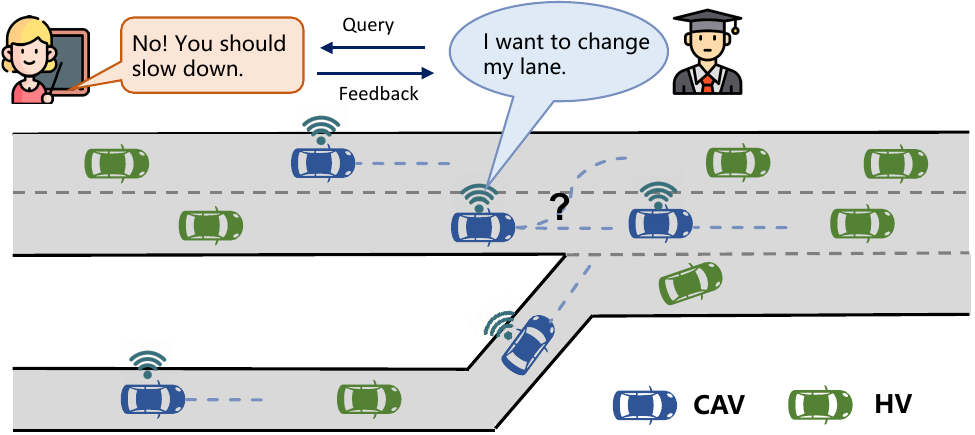}
  \caption{Illustration of the multi-CAV decision-making task with LLM guidance at the merge scenario. Each CAV agent will make decision with the consideration of reasoning and advice from LLM and update policy with knowledge distilled by LLM.
  }
  \label{fig:scenario}
\end{figure}

As shown in Fig.\ref{fig:scenario}, we use a ramp merging environment to train and test our method. Our experiments show that a small amount of high-quality teaching from the teacher significantly improves student policies' exploration efficiency, leading to rapid performance gains that eventually surpass the teacher's. Extensive experiments demonstrate that our teaching system effectively enhances CAV learning, outperforming baseline methods in learning efficiency and performance.

Our contributions are summarized as follows:
\begin{itemize}[leftmargin=*]
    \item We model the cooperative decision-making problem of CAVs as a MARL problem and utilize LLM to guide the exploration of CAVs.
    \item We propose LDPD, a language-driven policy distillation framework to facilitate rapid learning and efficient exploration for smaller policy networks.
    \item Through the teacher's guidance, the performance of the Student agents demonstrates significant advantages. Our method outperforms baseline approaches in terms of both learning efficiency and overall performance.
\end{itemize}

\section{Related Works}
\textbf{Cooperation Decision-Making of CAVs.}
In complex mixed-autonomy traffic environments, cooperative decision-making technologies can effectively improve the efficiency and safety of the CAV system. Numerous methods have been proposed to address vehicle coordination, including rule-based methods~\cite{jacobson1989real}, optimized-based methods~\cite{wang2024homogeneous}, and learning-based methods~\cite{li2024flexible}. Among them, learning-based methods are highly effective in simulating vehicle interaction dynamics and reasoning, while their lack of interpretability and limited generalization capabilities hinder their application in safety-critical scenarios.

\textbf{Large Language Model for Decision-making.}
Large Language Models (LLMs), known for their human-like reasoning and commonsense generalization, have recently demonstrated potential beyond traditional natural language processing tasks ~\cite{mees2023grounding, li2022competition}. 
In recent years, LLMs have been utilized as employed as end-to-end agents capable of autonomous decision-making.  
For cooperative driving tasks, Hu et al. proposed AgentsCoDriver~\cite{hu2024agentscodriver}, a LLM-driven lifelong learning framework to empower CAV decision-making. Meanwhile, CoDrivingLLM, proposed by Fang et al.~\cite{fang2024codrivingllm}, is designed to adapt different levels of Cooperative Driving Automation, which could effectively address the dilemma of cooperative scenarios. 
While LLMs have demonstrated remarkable potential, significant challenges remain. Their high computational demands and the complexity of real-time decision-making introduce latency issues, which limit their application in time-critical environments. 

\textbf{MARL and LLM-based MARL.}
Multi-Agent Reinforcement Learning (MARL) is an evolving research area that addresses systems composed of multiple interacting agents, such as robots, vehicles, and machines operating within a shared environment~\cite{chen2023deep, guo2024mappo,zhang2024multi}. 
However, MARL approaches often face challenges in scalability, inter-agent coordination, and exploration when applied to more dynamic and uncertain real-world scenarios. 
To address these limitations, integrating Large Language Models (LLMs) into MARL frameworks has emerged as a promising direction. LLM-based MARL approaches aim to harness LLMs' reasoning and knowledge capabilities to improve agents' decision-making processes ~\cite{sun2024llm}. 
Recent studies have explored LLMs' potential to boost the learning efficiency and performance of Reinforcement Learning (RL) agents. For example, in ~\cite{zhang2024large}, a semi-parametric RL agent was developed by integrating a Rememberer module that utilizes LLMs to retain long-term experiential memory. Similarly, ~\cite{zhou2023large} employed LLM-based instructions to train agents and distill prior knowledge, resulting in enhanced RL performance. Despite these advancements, LLMs' application in MARL remains underexplored. Current approaches have yet to fully capitalize on LLMs' ability to interpret environmental cues and negotiate high-level strategies and intentions among agents.

\section{Problem Formulation}
The cooperative decision-making problem for multiple CAVs can be modeled as a partially observable Markov decision process (POMDP)~\cite{spaan2012partially}. We define the POMDP using the tuple $\mathcal{M}_{\mathcal{G}} = (\mathcal{V}, S, [\mathcal{O}_i], [\mathcal{A}_i], \mathcal{P}, [r_i])$, where $\mathcal{V}$ represents the finite set of all controlled agents (CAVs), and $S$ denotes the state space encompassing all agents. $\mathcal{O}_i$ represents the observation space for each agent $i \in \mathcal{V}$, $\mathcal{A}_i$ denotes the action space, and $r_i$ is the reward associated with CAV $i$. The transition distribution is represented by $\mathcal{P}$.
At any given time, each agent $i$ receives an individual observation $\mathcal{O}_i$ and selects an action $a_i \in \mathcal{A}_i$ based on a policy $\pi_i : \mathcal{O}_i \times \mathcal{A}_i \to [0,1]$. 

\subsection{Observation Space}
The observation matrix for agent $i$, denoted as $\mathcal{O}_i$, is a matrix with dimensions $| \mathcal{N}_i | \times | \mathcal{F} |$, where $|\mathcal{N}_i|$ represents the number of observable vehicles for agent $i$, and $| \mathcal{F} |$ is the number of features used to describe a vehicle's state. The feature vector for vehicle $k$ is expressed as
\begin{equation}
\mathcal{F}_k = [x_k, y_k, v^x_k, v^y_k, \cos{\phi_k}, \sin{\phi_k}],
\end{equation}
where $x_k$ ,$y_k$ $v^x_k$ and $v^y_k$ are the longitudinal and lateral positions and speeds, respectively. $\phi_k$ is the vehicle's heading angle.
The overall observation space of the system is the combined observation of all CAVs, i.e., $\mathcal{O} =\mathcal{O}_{1}\times \mathcal{O}_{2}\times \cdots \times \mathcal{O}_{| \mathcal{V} |}$.

\subsection{Action Space}
Given that the strength of LLMs lies in their reasoning capabilities based on world knowledge rather than numerical computation, we design the decision actions of CAVs as discrete semantic decisions rather than direct vehicle control actions.
The action space $\mathcal{A}_i$ for agent $i$ is defined as a set of high-level control decisions, including $\{\textit{slow down}, \textit{cruise}, \textit{speed up}, \textit{change left}, \textit{change right}\}$. Once a high-level decision is selected, lower-level controllers generate the corresponding steering and throttle control signals to manage the CAVs' movements. The overall action space is the combination of actions from all CAVs, i.e., $\mathcal{A} =\mathcal{A}_{1}\times \mathcal{A}_{2}\times \cdots \times \mathcal{A}_{| \mathcal{V} |}$.

\subsection{Reward function}

In this paper, with the objective of ensuring that all agents pass through the merging area safely and efficiently, based on previous work~\cite{guo2024mappo},the reward of $i$th agent at the time step $t$ is defined as follows:
\begin{equation}
  {r}_{i,t}={\omega }_{c}{r}_{c}+{\omega }_{s}{r}_{s}+{\omega }_{h}{r}_{h}+{\omega }_{m}{r}_{m}.
\end{equation}
where ${\omega }_{c},{\omega }_{s},{\omega }_{h}$ and ${\omega }_{m}$ are positive weighting coefficients of collision reward ${r}_{c}$, stable-speed reward ${r}_{s}$, headway cost reward ${r}_{h}$ and merging cost reward ${r}_{m}$, respectively.

\section{Method}
In this section, the whole LDPD framework is first summarized. Then the teacher agent and student agent are introduced respectively.

\subsection{Framework Overview}
\begin{figure*}
  \centering
  \includegraphics[width=0.7\textwidth]{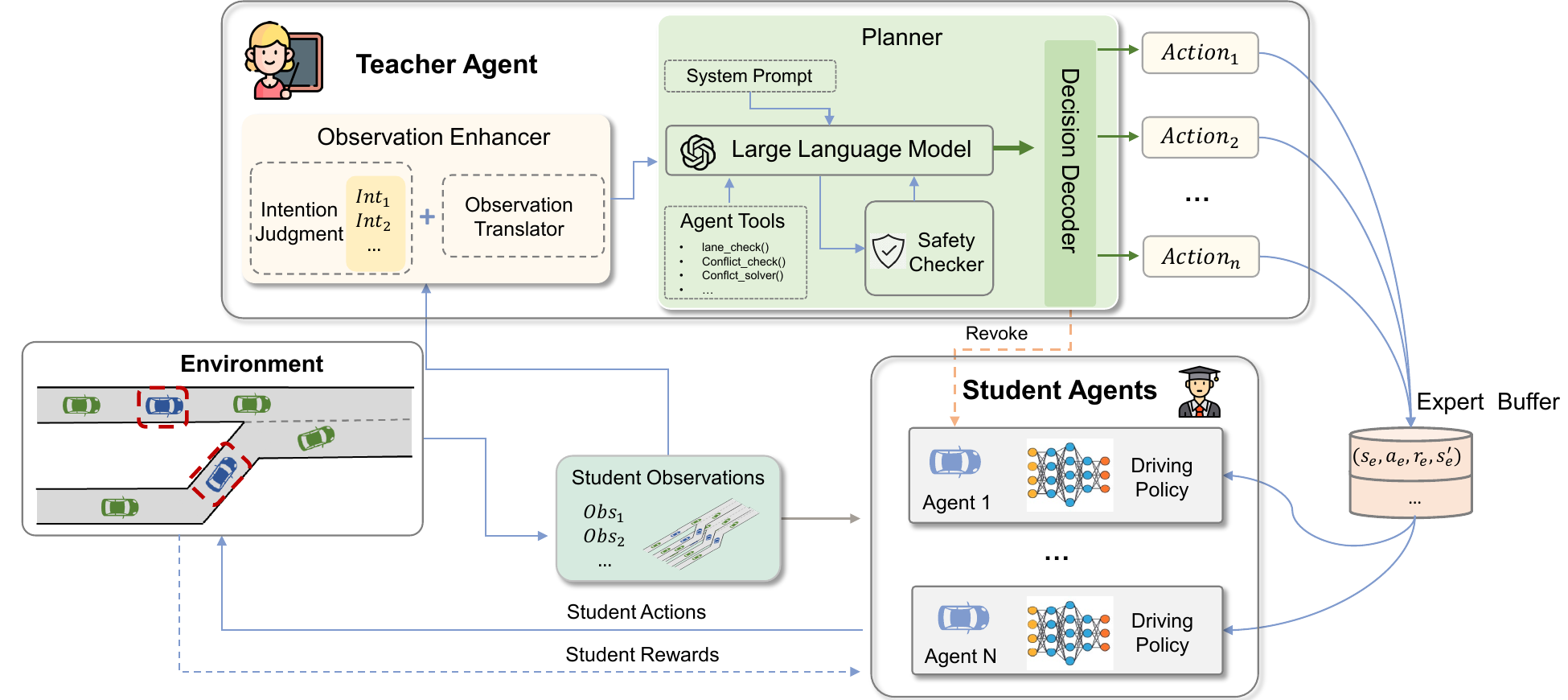}
  \caption{Illustration of the proposed LDPD framework, consisting of two components: a teacher agent powered by LLM and student agents modeled by MARL. The teacher agent processes observations from the student agents and, using LLM-based planning, outputs expert-level action recommendations. In the early exploration stage, the student agents accelerate the learning and updating of their policy networks by distilling knowledge from the teacher’s demonstrations.
  }
  \label{fig:framework}
\end{figure*}

The LDPD framework is designed to facilitate the learning and exploration process of multi-agent systems with distilled knowledge from LLM, especially in complex cooperative driving scenarios. 
 As shown in Fig.\ref{fig:framework}, the LDPD framework comprises two main components: the teacher agent and the student agents. The teacher agent facilitates efficient policy learning for multiple student agents through its action demonstrations. 
The student agents, composed of multiple small policy networks, form an agent group where each network is responsible for controlling an individual CAV. Throughout this process, the student agents efficiently distill and apply the knowledge from the LLM, enabling rapid learning and adaptation.

\subsection{Teacher Agent}
The teacher agent is the core of our entire framework, comprising the Observation Enhancer, Planner, and Safety Checker.

\subsubsection{Observation Enhancer}
This module is primarily responsible for collecting perception information from the CAV and enhancing the state information. The teacher first receives the observation $\mathcal{O}$ from the Student agents and then reorganizes this information.

Specifically, we categorize the lanes into three types: Lanes = \{lane\_ego, lane\_adj, lane\_conf\}, where lane\_ego, lane\_adj, and lane\_conf represent the ego lane, adjacent lanes, and conflict lanes, respectively. The ego lane is the lane in which the vehicle is currently driving, the adjacent lanes are those on either side of the ego lane (if they exist), and the conflict lanes are those intersecting with the ego lane. Similarly, based on the relationship between the vehicle's lane and the ego lane, we classify vehicle information into front vehicle (Veh\_f), rear vehicle (Veh\_r), surrounding vehicles (Veh\_s), and conflict vehicles (Veh\_c). From these classifications, we can summarize the surrounding environment information for each CAV.

Additionally, based on the state information of the CAV, we predict and infer the driving intentions of each CAV, including the desired target lane and intended driving behavior. At time \(t\), the driving intention information for CAV \(i\) is represented as \(In_{i,t} = \{lane'_{i,t}, behavior_{i,t}\}\), where \(lane'_{i,t}\) and \(behavior_{i,t}\) refer to the desired lane intention and the desired behavior intention, respectively.

Using the state and intention information, we generate a semantic driving scenario description for each CAV:
\begin{equation}
    Sce_i = Enhancer\{O_i, Veh_i, Lanes_i, In_i\}.
\end{equation}

\subsubsection{Core Planner}
The LLM-based Planner is the core of the teacher agent, integrating reasoning and decision-making actions within the LLM using the ReAct method~\cite{yao2022react}. We have designed a series of Agent tools to assist the LLM in making expert-level teacher decisions. These tools include lane querying, vehicle state prediction, conflict checking, conflict resolution, and more.

We use the conflict checking function as an example to introduce our tools:
After the LLM obtains the semantic description of the scenario, it will call the conflict checking tool $conflict\_check()$ before initiating reasoning to detect potential safety risks.
\begin{equation}
Risk = LLM\{conflict\_check(Sce_i)\}
\end{equation}

The conflict checking tool first retrieves all potential conflicts in the current environment and assesses the risk level based on the current states of the conflicting parties. 
Here, we utilize the time-to-conflict-point (TTCP) difference~\cite{fang2024codrivingllm} as a surrogate metric for the risk level, which is widely used to evaluate collision risks at potential conflict points between vehicles, calculated as follows:
\begin{equation}
\begin{aligned}
\Delta TTCP & = \begin{vmatrix} TTCP_i - TTCP_j \end{vmatrix} = \begin{vmatrix} \frac{d_{i}}{v_{i}} - \frac{d_{j}}{v_{j}} \end{vmatrix} \\
\end{aligned}
\label{ttcp}
\end{equation}
where $TTCP$ is the time to conflict point based on the vehicle's current distance to the conflict point $d$ and speed $v$.

By invoking the conflict checking module, we analyze the conflict information in the ramp merging scenario and convert this information into a semantic description that is provided to the LLM. Based on the scenario and conflict information, the LLM generates an initial decision:
\begin{equation}
\label{eq:output_ori}
A = LLM(Sce, conflict\_info)
\end{equation}

\subsubsection{Safety Checker}
After the initial decision action set is generated, it enters a Safety Checker for further safety verification to ensure the accuracy of the teacher’s decision results. The Safety Checker first re-evaluates conflicts and intentions based on the previously predicted intentions and the LLM’s decision output. It begins by acquiring the intention trajectories of all CAVs and their surrounding vehicles for the next ${T}_{n}$ steps.

Firstly, the formula is designed to calculate the priority score of each CAV~\cite{guo2024mappo}:
\begin{equation}
{p}_{i}={\alpha }_{1}{p}_{m}+{\alpha }_{2}{p}_{e}+{\alpha }_{3}{p}_{h}+{\sigma }_{i}.
\label{eq:priority}
\end{equation}
where ${p}_{i}$ denotes the priority of agent $i$, ${\alpha }_{1}$, ${\alpha }_{2}$, and ${\alpha }_{3}$ are the positive weighting parameters for the merging metric ${p}_{m}$, merging-end metric ${p}_{e}$, and time headway metric ${p}_{h}$, respectively. A random variable ${\sigma }_{i}\sim\mathcal{N}(0,0.001)$ is added as a small noise to avoid the issue of identical priority values.

Specifically, ${p}_{m}$ is defined as:
\begin{equation}\label{Eq:pm}
{p}_{m}=\left\{\begin{matrix}
0.5,& \text{If on merge lane,}\\
0,& \text{Otherwise.}
\end{matrix}\right.
\end{equation}
which indicates that vehicles on the merge lane should be prioritized over those on the through lane due to the urgency of the merging task.

Next, since merging vehicles closer to the end of the merge lane should be given higher priority due to a greater risk of collision and deadlocks, the merge-end priority value is calculated as follows:
\begin{equation}\label{Eq:pe}
{p}_{e}=\left\{\begin{matrix}
\frac{x}{L},& \text{If on merge lane,}\\
0,& \text{Otherwise.}
\end{matrix}\right.
\end{equation}
where $L$ is the total length of the merging lane, and $x$ is the position of the ego vehicle on the ramp.

Finally, we define the time-headway priority as:
\begin{equation}\label{Eq:ph}
{p}_{h}=-\log_{}\left(\frac{{d}_{headway}}{{t}_{h}{v}_{t}}\right)
\end{equation}
indicating that vehicles with smaller time-headway are more dangerous than others.

With the priority scores of CAVs calculated above, at time step $t$, the safety checker first produces a priority list ${P}_{t}$, which consists of a list of priority scores and their corresponding ego vehicles. Then, the safety checker sequentially examines the future trajectory of the vehicles in the list to determine if there are any conflicts between vehicles.
We use the Intelligent Driver Model (IDM)~\cite{treiber2000congested} to predict the longitudinal acceleration of HVs, based on their current speed and distance headway, and we utilize the MOBIL lane change model~\cite{kesting2007general} to predict the lateral behavior of HVs.

If a collision is detected, the Safety Checker module generates a new safety decision to correct the actions of the Planner. The correction process is as follows:
\begin{equation} \label{Eq: action_correct}
{a}_{t}^{'}=\arg\max_{{a}_{t}\in {A}_{available}}({\min_{k\in {T}_{n}}{{d}_{sm,k}}}).
\end{equation}
where ${A}_{available}$ is a set of available actions at time step $k$ for the selected agent, and ${d}_{sm,k}$ is the safety margin at prediction time step $k$.
The safety margin is defined as follows:
\begin{equation}
\scriptsize
{d}_{sm,k}=\left\{\begin{matrix}
\min{\left \{|{P}_{{v}_{t},k}-{P}_{{v}_{e},k}|,|{P}_{{v}_{c},k}-{P}_{{v}_{e},k}| \right \}},& \text{If change lane,}\\
{P}_{{v}_{pre},k}-{P}_{{v}_{e},k},& \text{Otherwise.}
\end{matrix}\right.
\end{equation}
where ${P}_{{v}_{t},k}$ and ${P}_{{v}_{c},k}$ denote the longitudinal positions of the preceding and following vehicles relative to the ego vehicle on both the target and current lanes, respectively. ${P}_{{v}_{pre},k}$ represents the position of the preceding vehicle at time step $k$, and ${P}_{{v}_{e},k}$ is the position of the ego vehicle. 

Finally, the teacher agent decodes and outputs the decision results, which include the set of actions guiding all CAVs: $A^*=\{a_1^*,a_2^*,\dots,a_n^*\}$, where
\begin{equation}
A^* = SafetyChecker(A).
\end{equation}

After the entire decision-making process of the teacher is completed, the decision results are stored in the Expert Buffer $\mathcal{D}$ for subsequent training of the Student Agents.

\begin{algorithm}
\SetAlFnt{\scriptsize}
\SetAlgoNlRelativeSize{-1}
\SetKwInOut{Parameter}{Inputs}
\SetKwInOut{Output}{Output}
\caption{Teacher Agent}
\LinesNumbered 
\label{alg:llm_teacher}
\SetAlgoNlRelativeSize{-1}
\SetAlgoVlined
\Parameter{Student observations $\mathcal{O}^{t}$, Student list $\mathcal{C}$}
\Output{Decisions set of teacher $\mathcal{A}^{t+1}_{Te}$}
\SetAlgoNlRelativeSize{-1}
\hrule
Initialize scenario description buffer $\boldsymbol{Sce}$;\\
\For{$i \in \mathcal{C}$}
{
    Generate the scenario description $sce_{i}$ of student $i$ based on current states $\mathcal{S}^{t}_{i}$;\\
    Add $sce_{i}$ to buffer $\boldsymbol{Sce} \leftarrow \boldsymbol{Sce}.\text{add}(sce_{i})$;
}
\For{$i \in \mathcal{C}$}
{
    Obtain the action $a_i^{t+1}$ of LLM by \eqref{eq:output_ori};\\
    Check the safety of action $a_i^{t+1}$ and solve the conflict with Safety Checker;\\
    Obtain the optimal action $a_i^{*t+1}$;\\
    Add decision to buffer $\mathcal{A}^{t+1}_{Te} \leftarrow \mathcal{A}^{t+1}_{Te}.\text{add}(a_i^{*t+1})$;
}
\end{algorithm}

\subsection{Student Agents}
\begin{figure*}
  \centering
  \includegraphics[width=0.7\textwidth]{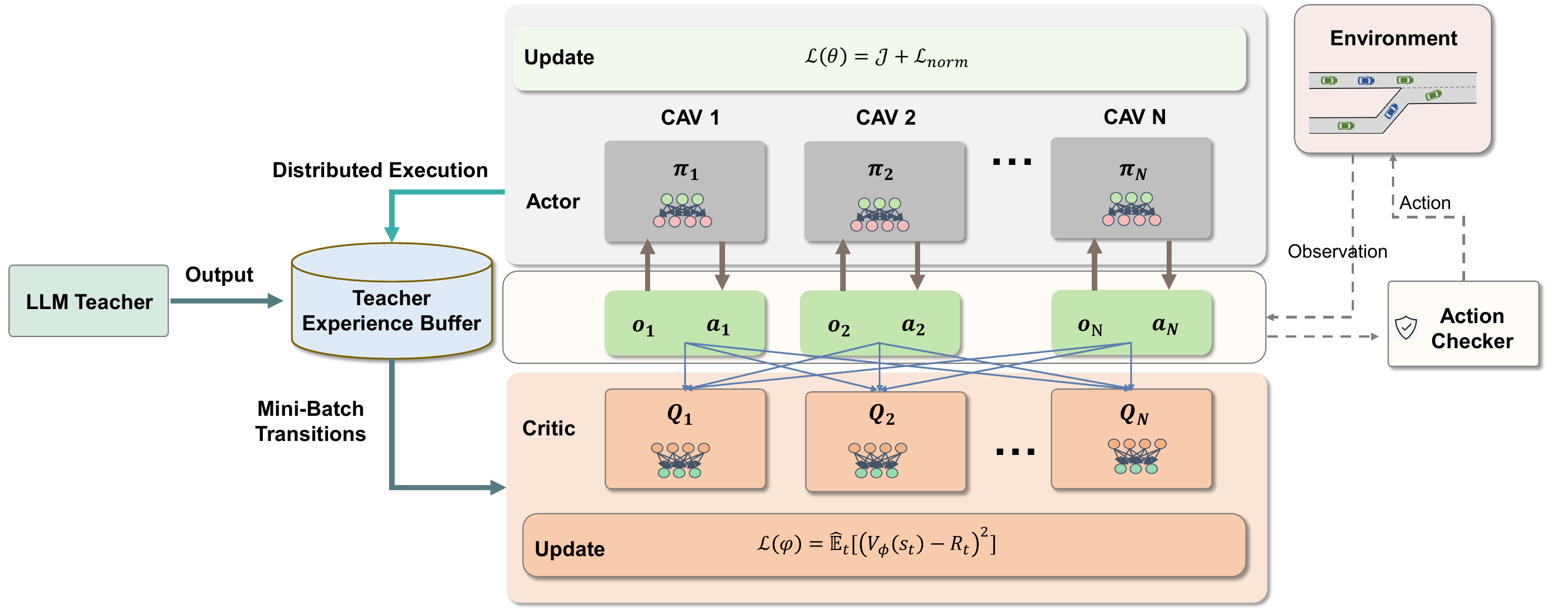}
  \caption{Illustration of policy updating and optimal action generation by student agents. The policy gradients of all student agents are updated using the actor-critic method. Expert-level demonstrations from the teacher agent are stored in the experience buffer. During the early exploration stage, the teacher’s demonstrations are sampled and used to update the students’ policies. Afterward, the student agents proceed with independent exploration.}
  \label{fig:student_update}
\end{figure*}

The Student Agents are composed of multiple small policy networks that form an agent group. In our scenario, each agent represents a single CAV, with its policy network functioning as the CAV's controller. We model these Student Agents using the Actor-Critic framework and train them under a CTDE approach. This means that each CAV consists of two models: the actor and the critic. Each actor undergoes centralized training via a centralized critic while making distributed, independent action decisions.

For student agent $i$, let ${\pi}_{\theta}^{i}$ represent the actor network approximating the policy, and ${V}_{\phi }^{i}$ denote the critic network approximating the value function, where $\theta$ and $\phi$ are the parameters of the actor and critic networks, respectively.

The policy network of each CAV can be updated as follows:
\begin{equation}\label{eqn:advantagepolicygradient_MARL}
{\mathcal{J}}_{i}(\theta ) = E_{\pi^i_\theta} \left[ \log \pi^i_\theta (a_{i, t}|s_{i, t})
 A^t_i \right],
\end{equation}
where $A^t_i = r_{i,t} + \gamma V_{\phi_i}(s_{i,t+1}) - V_{\phi_i}(s_{i,t})$ is the advantage function  and $V_{\phi_i}(s_{i,t})$ is the state value function.

To ensure the effectiveness of learning and the independence of the agent's policy, we introduce a regularization term in the loss function:
\begin{equation}
    \mathcal{L}_{i_{\text{norm}}} = \lambda\mathbb{E}_{s\sim\pi_{\theta}}\mathcal{H}\left(\pi_T(\cdot|s)|| \pi_\theta^i(\cdot|s)\right)
\end{equation}
where the regularization term $\mathcal{H}\left(\pi_T(\cdot|s)||\pi_\theta^i(\cdot|s)\right)$, describes the difference between the teacher policy $\pi_T$ and the student policy $\pi_\theta^i$, calculated using the Kullback-Leibler (KL) divergence between these two policies.
$\lambda$ is an annealing parameter we introduce to control the student agent's dependence on the teacher agent. When $\lambda$ is set to zero, the student's learning process simplifies to a standard RL process, unaffected by the teacher agent. At the initial stage of training, we initialize the annealing parameter $\lambda$ with a large value. This setup ensures that the student agent focuses more on the guidance provided by the LLM-based teacher agent, aiming to align its policy with the teacher's policy. As training progresses, we gradually decay $\lambda$ linearly, allowing the student agent to shift its focus towards maximizing its expected return. By reducing the influence of the teacher's guidance, the student agent becomes more independent in its decision-making process and emphasizes its own learning strategy.

Therefore, for each agent $i$, the actor ultimately updates the policy by minimizing the following objective function:
\begin{equation}
\begin{aligned}
    & \mathcal{L}_i(\theta) =  {\mathcal{J}}_{i}(\theta ) + \mathcal{L}_{i_{\text{norm}}} \\
    &= {\mathcal{J}}_{i}(\theta )  + \lambda\mathbb{E}_{s\sim\pi_{\theta}}\mathcal{H}\left(\pi_T(\cdot|s) \| \pi^i_\theta(\cdot|s)\right)
\label{eq:actor_loss}
\end{aligned}
\end{equation}

For each student agent $i$, the critic network is trained using a stochastic gradient descent (SGD) method to minimize the value loss, defined by the following loss function:
\begin{equation}
  \mathcal{L}^i(\phi) = \hat{\mathbb{E}_{t}}[({V}^i_{\phi}({s}_{i,t}) - {R}_{i,t})^2],
\end{equation}
where ${R}_{t}$ is the cumulative return, and ${V}_{\phi}({s}_{i,t})$ is the current value estimate of agent $i$. The schematic diagram is shown in Fig.\ref{fig:student_update}.

Additionally, during the actual exploration process, the student agents are allowed to leverage the teacher agent's decision-making tools to further enhance the quality of their decisions. 
The teaching process will be executed for a couple of iterations $E_{te}$ to improve the learning efficiency of students. After that, the student agents will continue to explore by themselves to facilitate policy diversity.
The learning process of the student agents is summarized in Algorithm \ref{alg:student_learing}.

\begin{algorithm}
\SetAlFnt{\scriptsize} 
\SetAlgoNlRelativeSize{-1}
\SetKwInOut{Parameter}{Inputs}
\SetKwInOut{Output}{Output}
\caption{The Procedure of Student Agents Group's Policy Learning}
\label{alg:student_learing}
\LinesNumbered 
\SetAlgoNlRelativeSize{-1}
\SetAlgoVlined
\Parameter{Teacher agent, students' policies $\{\pi_1, \cdots, \pi_k\}$, number of teaching iterations $E_{te}$, number of self-learning iterations $E_{self}$, maximum timestep $T_{max}$
}
\Output{$\theta$}
\hrule
\CommentSty{Phase 1: Policy Distillation}\\
\For{$episode = 1$ to $E_{te}$}
{
    \For{$t=1$ to $T_{max}$}
    {
    Collect rollouts following the student's initial policy;\\
    Get action of the teacher according to Alg. \ref{alg:llm_teacher};\\
    Update $\mathcal{D}_i \leftarrow (\textbf{o}, a^{*}_{i,t}, r_{i,t}, \textbf{o}^{\prime}_{i,t})$;\\
    \For{ Student $i=1 \in \mathcal{V}$} 
    {
        $\theta^i \leftarrow \theta^i - \alpha \nabla_{\theta^i}(\mathcal{J}_{i}(\theta^i )  + \lambda_i\mathbb{E}_{s}\mathcal{H}\left(\pi_T(\cdot|s) || \pi_\theta^i(\cdot|s)\right))$;\\
    }
}
}
\CommentSty{Phase 2: Self learning}\\
\For{$episode = 1$ to $E_{self}$}
{
{\bf Initialize} buffer of student agent $\mathcal{D} \leftarrow \emptyset$;\\
    \For{$Student \ i = 1 \in \mathcal{V} $}
    {
        Get observation $o_{i,t}$;\\ 
        Update action $a_i=\pi_{\theta_i}(o_{i,t})$;\\
        Update $\mathcal{D}_i \leftarrow (\textbf{x}_{i,t}, a^{*}_{i,t}, r_{i,t}, \textbf{x}^{\prime}_{i,t})$;\\
    }
    \For{ Student $i=1 \in \mathcal{V}$}
    {
        Update critic network: $\phi^i \leftarrow \phi^i - \alpha \nabla_{\phi^i}(\mathcal{J}_{i}(\phi^i)) $;\\
        Update actor network:
        $\theta^i \leftarrow \theta^i - \beta \nabla_{\theta^i}(\mathcal{J}_{i}(\theta^i)) $;\\
    }
}
\end{algorithm}

\section{Experiment}

\subsection{Experimental Setups}
\begin{figure}
  \centering
  \includegraphics[width=0.4\textwidth]{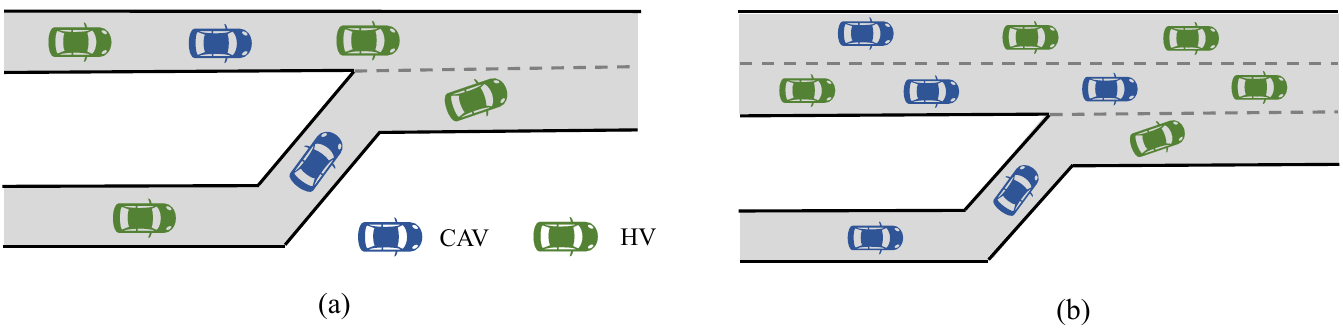}
  \caption{The two scenarios used in our experiments, (a) scenario \#1, (b) scenario \#2.
  }
  \label{fig:scenario_setting}
\end{figure}

Our environment simulator is developed based on highway-env~\cite{highway-env}, an open-source and flexible simulation platform for autonomous driving.
We evaluate our method in a ramp merging scenario, which is one of the most complex traffic scenarios in the real world. CAVs on the ramp must accurately assess safe distances and merging timing, while coordinating with other CAVs on the ramp and main road to ensure safe merging. As shown in Fig.\ref{fig:scenario_setting}, we design two ramp merging scenarios, and for each scenario, we define three different traffic densities (easy, medium, and hard) to simulate a range of real-world conditions. The three traffic densities generate between 2-4, 4-6, and 6-8 CAVs and HVs randomly. In these scenarios, all CAVs are controlled by our model, while HVs are managed by IDM and MOBIL to simulate basic interaction behaviors. We select MAACKTR~\cite{wu2017scalable}, MAPPO~\cite{yu2022surprising}, and MAA2C as baselines to compare against the performance of our method.

\subsection{Implementation details} 
For the MARL algorithm, we set the discount factor $\gamma = 0.99$, exploration rate $\epsilon = 1e-5$, and advantage decay rate $\alpha = 0.99$. The learning rates for both the actor and critic networks are set to $5e^{-4}$. The priority coefficients $\alpha_1$, $\alpha_2$, and $\alpha_3$ are uniformly set to 1.
The algorithms are evaluated by running 3 episodes every 200 training episodes, with all models being trained over 20,000 episodes in total.
GPT-4o is employed as our base LLM model, facilitating high-level logical reasoning and decision-making for the teacher agent. The experiences derived from teacher agent are utilized for the initial 2,000 episodes of training. After this phase, student agents are allowed to explore independently.
To ensure reliability, all scenarios under different settings are replicated three times with distinct random seeds to obtain consistent results.


\subsection{Performance Analysis}

\subsubsection{Overall Performance}
\begin{figure*}
  \centering
  \includegraphics[width=0.8\textwidth]{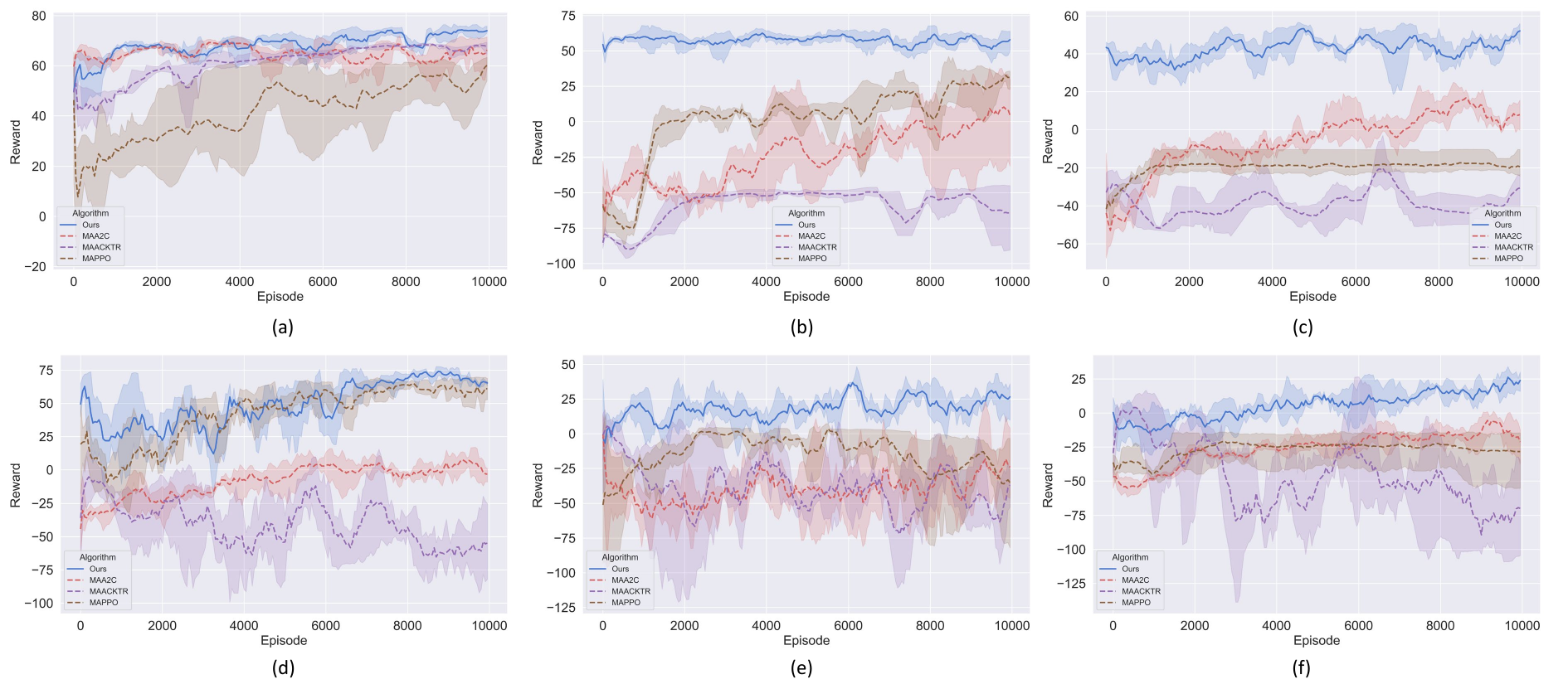}
  \caption{The training results under different experimental settings, (a), (b) and (c) are experimental results with easy, medium and hard mode at scenario \#1, respectively; (d), (e) and (f) are experimental results with easy, medium and hard mode at scenario \#2, respectively.
  }
  \label{fig:training_result}
\end{figure*}

The training curves for all methods under different conditions and scenarios are shown in Fig.\ref{fig:training_result}. As observed, in all experimental scenarios, our method (LEMAA2C) consistently outperforms the MARL baselines. Furthermore, they demonstrate significantly higher exploration efficiency in the early stages of training, which effectively boosts the training efficiency of the student agents. This strongly validates the effectiveness of our approach.
Meanwhile, we also compare our method with some LLM-based approaches. As shown in Table \ref{tab:llm_model_performance}, our method exhibits the best performance across various tests, owing to its stronger exploration capability in the later stages of training.

\begin{table}[!htbp]
\centering
\caption{Performance comparison of different models under varying difficulty levels.}
\label{tab:llm_model_performance}
\begin{tabular}{lcccc}
\toprule
\textbf{Scenario} & \textbf{Model} & \textbf{Easy} & \textbf{Medium} & \textbf{Hard} \\
\midrule
\multirow{4}{*}{Scenario 1} & GPT-3.5~\cite{openai_introducing_2023} & 91\% & 69\% & 23\% \\
                            & GPT-4o~\cite{achiam2023gpt} & 98\% & 82\% & 58\% \\
                            & DiLu~\cite{wen2023dilu} & 96\% & 77\% & 60\% \\
                            & Ours & \textbf{100}\% & \textbf{96}\% & \textbf{87}\% \\
\midrule
\multirow{4}{*}{Scenario 2} & GPT-3.5 & 92\% & 72\% & 31\% \\
                            & GPT-4o & \textbf{100}\% & 78\% & 30\% \\
                            & DiLu & 99\% & 82\% & 34\% \\
                            & Ours & \textbf{100}\% & \textbf{81}\% & \textbf{67}\% \\
\bottomrule
\end{tabular}
\end{table}

\subsubsection{Safety Performance}
The model’s overall performance is tested under various experimental conditions, as shown in Tab.\ref{tab:performance_all_methods}. 
Post Encroachment Time (PET), a key safety metric that quantifies the interaction and risk level between two vehicles, is used as one of the primary evaluation criteria.
It is evident that our method consistently achieves the lowest collision rate across all tests. We also evaluate the safety performance of the teacher agent under different conditions and compare it with the student agents, as shown in Tab. \ref{tab:performance_teacher}. 
In most scenarios, LAMAA2C outperforms LLM in the final performance, demonstrating that our method possesses strong exploration capabilities that further enhance performance. We believe this is due to the fact that as training progresses, the student model is able to integrate its own exploration and learning through policy updates more freely, in addition to receiving guidance from the teacher.

\begin{table*}[!htbp]
\centering
\caption{Performance across scenarios for various methods under different difficulty settings}
\label{tab:performance_all_methods}
\begin{tabular}{cccccccccc}
\toprule
\textbf{Difficulty} & \textbf{Model} & \multicolumn{2}{c}{\textbf{Evaluation Reward}} & \multicolumn{2}{c}{\textbf{Collision Rate}} & \multicolumn{2}{c}{\textbf{Average Speed}} & \multicolumn{2}{c}{\textbf{Average PET}} \\
\cmidrule(r){3-4} \cmidrule(lr){5-6} \cmidrule(lr){7-8} \cmidrule(l){9-10}
 &  & Sce-1 & Sce-2 & Sce-1 & Sce-2 & Sce-1 & Sce-2 & Sce-1 & Sce-2 \\
\midrule
\multirow{4}{*}{Simple} 
 & MAA2C & 67.52 & 53.03 & 0.01 & 0.02 & 28.14 & 25.26 & 18.68 & 19.97 \\
 & MAPPO & 30.31 & 58.63 & 0.17 & 0.09 & 23.65 & 28.79 & 13.29 & 19.61 \\
 & MAACKTR & 13.96 & -29.92 & 0.27 & 0.51 & 25.51 & 25.46 & 11.08 & 7.97 \\
 & Ours & \textbf{80.44} & \textbf{75.15} & \textbf{0.00} & \textbf{0.00} & 27.14 & 27.70 & 19.07 & 17.67 \\
\midrule
\multirow{4}{*}{Medium} 
 & MAA2C & 16.85 & -1.58 & 0.23 & 0.30 & 23.96 & 25.17 & 8.76 & 7.67 \\
 & MAPPO & 47.68 & 2.35 & 0.09 & 0.34 & 21.76 & 22.65 & 9.33 & 12.59 \\
 & MAACKTR & 0.07 & -14.99 & 0.44 & 0.43 & 23.48 & 25.64 & 7.48 & 8.27 \\
 & Ours & \textbf{56.91} & \textbf{35.97} & \textbf{0.04} & \textbf{0.19} & 24.11 & 26.07 & 7.16 & 7.40 \\
\midrule
\multirow{4}{*}{Hard} 
 & MAA2C & 17.76 & -0.64 & 0.30 & 0.53 & 22.29 & 22.63 & 7.01 & 4.37 \\
 & MAPPO & -37.03 & -35.55 & 0.81 & 0.96 & 25.29 & 22.66 & 4.05 & 4.18 \\
 & MAACKTR & -12.47 & -25.11 & 0.65 & 0.51 & 20.65 & 25.50 & 5.74 & 9.28 \\
 & Ours & \textbf{36.87} & \textbf{14.33} & \textbf{0.13} & \textbf{0.33} & 21.98 & 23.11 & 5.88 & 4.00 \\
\bottomrule
\end{tabular}
\end{table*}

\begin{table}[!htbp]
\centering
\caption{Performance of teacher and students under varying difficulty levels.}
\label{tab:performance_teacher}
\begin{tabular}{lccccc}
\toprule
\textbf{Scenario} & \textbf{Role} & \textbf{Easy} & \textbf{Medium} & \textbf{Hard} \\
\midrule
Scenario 1 & Teacher & 98\% & 82\% & 58\% \\
           & Students & \textbf{100}\% & \textbf{96}\% & \textbf{87}\% \\
\midrule
Scenario 2 & Teacher & \textbf{100}\% & 78\% & 30\% \\
           & Students & \textbf{100}\% & \textbf{81}\% & \textbf{67}\% \\
\bottomrule
\end{tabular}
\end{table}

\subsubsection{Efficiency Performance}
The average speed of all CAVs, shown in Tab.\ref{tab:performance_all_methods},  is analyzed to assess the efficiency of cooperative driving across different methods. We observe that as the difficulty of tasks increases, the average speed of the CAVs decreases, which likely represents a compromise made to ensure safety over efficiency. However, compared to other baseline methods, such as MAPPO and MAACKTR, our approach not only maintains better safety performance but also ensures driving efficiency. This demonstrates that our method effectively balances driving speed with safety.

\subsubsection{Cross-Validation}
To evaluate the adaptability and generalization capabilities of the learned strategies, we test the models in task environments beyond their training conditions. For example, models trained under the hard mode are now tested on easy and medium mode tasks, with results shown in Tab.\ref{tab:cross_test}. We observe that models trained in hard mode not only outperform the original medium mode models in terms of safety and efficiency but also excel in other scenarios. This demonstrates the exceptional adaptability of our model.

\begin{table}[!htbp]
\centering
\caption{Performance comparison of models across different scenarios and difficulty levels.}
\label{tab:cross_test}
\resizebox{1\linewidth}{!}{
\begin{tabular}{llcccc}
\toprule
\textbf{Original Level} & \textbf{Applied Level} & \multicolumn{2}{c}{\textbf{Evaluation Reward}} & \multicolumn{2}{c}{\textbf{Collision Rate}} \\
\cmidrule(r){3-4} \cmidrule(l){5-6}
 &  & Sce-1 & Sce-2 & Sce-1 & Sce-2 \\
\midrule
\multirow{2}{*}{HARD} & Simple & 62.66 & 46.18 & 0.02 & 0.13 \\
                      & Medium & 51.16 & 70.78 & 0.07 & 0.00 \\
\midrule
MEDIUM & Simple & 69.31 & 61.64 & 0.00 & 0.02 \\
\bottomrule
\end{tabular}
}
\end{table}

\section{Conclusions}
To enhance exploration and learning efficiency in cooperative decision-making for CAVs, we introduced LDPD, a language-driven policy distillation framework to facilitate MARL agents learning. Within this framework, the language model, as the teacher, makes complex reasoning decisions based on observations from student agents and stores these decision-making experiences. Student agents learn the teacher’s prior knowledge through gradient updates in their policy networks. Extensive experiments demonstrate that distilled information from the teacher significantly enhances the exploration efficiency of student agents. Compared to baseline methods, there is a notable overall improvement in the agents' performance, eventually surpassing the expert decision-making level of the teacher. In the future, we plan to extend our method to test additional MARL algorithms and apply it across a broader range of decision-making scenarios. 

\bibliographystyle{IEEEtran}  
\bibliography{reference}  

\vfill

\end{document}